# Deep Learning for Technical Document Classification

Shuo Jiang, Jie Hu, Christopher L. Magee, Jianxi Luo

*Abstract*— **In large technology companies, the requirements for managing and organizing technical documents created by engineers and managers have increased dramatically in recent years, which has led to a higher demand for more scalable, accurate, and automated document classification. Prior studies have only focused on processing text for classification, whereas technical documents often contain multimodal information. To leverage multimodal information for document classification to improve the model performance, this paper presents a novel multimodal deep learning architecture, TechDoc, which utilizes three types of information, including natural language texts and descriptive images within documents and the associations among the documents. The architecture synthesizes the convolutional neural network, recurrent neural network, and graph neural network through an integrated training process. We applied the architecture to a large multimodal technical document database and trained the model for classifying documents based on the hierarchical International Patent Classification system. Our results show that TechDoc presents a greater classification accuracy than the unimodal methods and other state-of-the-art benchmarks. The trained model can potentially be scaled to millions of real-world multimodal technical documents, which is useful for data and knowledge management in large technology companies and organizations.**

## I. Introduction

ENGINEERING processes involve significant technical and organizational knowledge and comprise a sequence of activities, such as design, analysis, and manufacturing [1]. During these engineering activities, a large amount of data and knowledge is generated and stored in various types of technical documents, such as technical reports, emails, papers, and patents [2], [3]. Prior studies have reported that engineers spend two-thirds of their time communicating to obtain a related document input for their work and make decisions based on such materials [4]. It is widely believed that 20% of engineering information can be extracted from a database comprising numeric data only, and the remaining 80% is hidden in the documents [5]–[7]. Feldman et al. [8] similarly asserted that 80% of explicit knowledge in companies can be found in their documents. With an increase in the scale and complexity of engineering activities, technical documents are being created at a greater pace than before [9].

A well-organized technical document classification enables engineers to retrieve and reuse documents more easily. However, a continuously increasing volume of technical documents requires engineers to spend much more time managing them than before. The label assignment and categorization of technical documents are human labor intensive, expensive, and time consuming. Because these documents are usually lengthy and full of complicated technical terminologies, it is also difficult to find specific experts to handle them. Specialized individual experts with limited knowledge and cognitive capacity might not be able to accurately determine the labels or categories of specific documents in a wide (many diverse classes) and deep (multi-level hierarchy) classification system. Therefore, we turn to artificial intelligence for reducing the time and cost and ensuring an accurate classification of technical documents.

Several prior studies have explored the use of machine learning algorithms to automatically classify technical documents for knowledge management [10]. Among them, several traditional methods, such as the K-nearest neighbors (KNN) and support vector machine (SVM), are not scalable and are incapable of classifying documents that large engineering companies such as Boeing or General Motors need to manage. Recent deep learning-based approaches have demonstrated the ability of a scalable classification on large engineering document datasets [11]. They focused on textual information and applied various NLP techniques to develop automated classifiers, such as recurrent neural networks (RNN), long short-term memory (LSTM) networks, and specific pre-trained models. However, the performance of current systems is insufficiently reliable for real-world applications. For example, Li et al. [12] reported that their CNN-based classifier achieved a precision of 73% and an F1 score of 42% on their curated dataset of two million patent documents.

Technical documents normally contain both text and images [13]. Ullman et al. studied the importance of technical drawings in the engineering design process [14]. In a technical document, visual information often plays important roles in presenting its novelty [15], [16]. In recent years, the data science community has explored multimodal deep learning that can utilize, process, and relate information from multiple modalities, and reported the superiority of a multimodal model over unimodal models for various tasks [17], [18], which

Shuo Jiang was with the Institute for Data, Systems, and Society, Massachusetts Institute of Technology, Cambridge 02139, USA. He is also with the School of Mechanical Engineering, Shanghai Jiao Tong University, Shanghai 200240, China (e-mail: jsmech@sjtu.edu.cn).
Jie Hu is with the School of Mechanical Engineering, Shanghai Jiao Tong University, Shanghai 200240, China (e-mail: hujie@sjtu.edu.cn).
Christopher L. Magee is with the Institute for Data, Systems, and Society, Massachusetts Institute of Technology, Cambridge 02139, USA (e-mail: cmagee@mit.edu).
Jianxi Luo is with the Data-Driven Innovation Lab, Singapore University of Technology and Design, 487372, Singapore (e-mail: luo@sutd.edu.sg).

presents new opportunities to improve the technical document classification. Also, technical documents are often associated with one another, via inter-document references or citations, indicating their coupling and embeddedness in a greater nearly-decomposable knowledge system [19]. Recent advanced graph neural networks enable us to learn such relational information and classify individual nodes into several pre-defined categories [20]. In addition, the technology knowledge space is a natural complex system and constitutes many knowledge categories and sub-categories corresponding to different technology fields [21]. When considering the classification of a large number and diversity of documents, a hierarchical classification system is required to assign documents into multi-layer categories, which has not been supported by current technical document classifiers yet.

In this work, we propose a multimodal deep learning-based model, TechDoc, for the accurate hierarchical classification of technical documents. Our aimed contribution is for the engineering management community with a focus on engineering document management, especially for those large engineering companies. Engineering documents are normally multimodal, and their classification needs to be hierarchical. Relevant automated classification methods that specifically address such requirements do not exist. Thus, this work is expected to bridge the gap between the up-to-date multimodal deep learning techniques and engineering document management.

Our TechDoc model utilizes three types of information (i.e., text, image and network) of engineering documents for the automated classification by synthesizing the convolutional neural network (CNN), recurrent neural network (RNN) and graph neural networks (GNN). To illustrate the proposed method, we applied it to a benchmark patent dataset and trained the model to classify technical documents based on the hierarchical International Patent Classification (IPC) system as the evaluation case study, which shows better performances than other existing classification methods. In addition, as far as we know, this study is the first effort to utilize and synthesize intrinsic information within technical documents and the associations among the documents to automate hierarchical technical document classification. Taken together, this research contributes to the growing literature on engineering knowledge management [22]–[24], patent analysis [25]–[29], and data-driven engineering applications [30]–[33].

This paper is organized as follows. In Section II, we briefly review the relevant literature about document classification and multimodal deep learning. Section III introduces the proposed TechDoc model in detail. Section IV presents a case study on a patent document dataset followed by a discussion on applications of the TechDoc in Section V. Finally, Section VI concludes the paper and discusses the limitations and opportunities for future research.

II. LITERATURE REVIEW AND BACKGROUND

*A. Document classification*

Document classification is a fundamental task in NLP and text mining, and to date, a wide variety of algorithms have exhibited significant progress. Traditional document classification approaches represent text with sparse lexical features, such as term frequency-inverse document frequency (TF-IDF) and N-grams, and then use a linear model (e.g., logistic regression) or kernel methods (e.g., SVM) based on these representations [34], [35]. In recent years, the development of high-performance computing has enabled us to take advantage of various deep learning methods and end-to-end training and learning, including CNNs [36], RNNs [37], capsule neural networks [30], and transformers [39]. For example, Joulin et al. [40] proposed a simple but efficient model called FastText, which views text as a bag of words and then passes them through one or more multilayer perceptrons for classification. Lai et al. [41] proposed a recurrent convolutional neural network (RCNN) for text classification without human-designed features. This model applied a recurrent structure to extract long-range contextual dependence when learning representations. In addition, Yang et al. [42] proposed a hierarchical attention network for document classification. In this model, the hierarchical structure mirrors the natural hierarchical structure of documents, and attention mechanisms are applied at both word- and sentence-level structures, enabling it to differentially attend to less and more important content when learning document representations. However, these models only use natural language data as the presentation of documents, and they are usually trained on a general document corpus that often involves a wide range of non-engineering topics.

Specifically, a few studies related to technical document classification have already existed in the engineering field. For example, Caldas et al. [10] described a document classification method based on a hierarchical structure from the Construction Specifications Institute. They used TF-IDF to represent the text and trained an SVM classifier to categorize the documents. The experiments were conducted using a dataset of 3,030 documents. Similarly, Chagheri et al. [43] used an SVM algorithm to train a classifier that helped Continew Co. classify and manage technical documents. Their model was trained and evaluated on a small set of 800 documents. These initial studies used traditional non-scalable machine learning techniques and were illustrated with small document sets.

Patent documents represent typical and complex engineering design documents. Prior studies have focused on patent document classification utilizing various machine learning and NLP techniques. For example, Fall et al. [44] and Tikk et al. [45] separately presented several basic classifiers on the World Intellectual Property Organization (WIPO) dataset, including NB, KNN, and SVM. The CLEF-IP tracks included a patent classification task [46], [47], which provided a dataset of more than 1 million patents as the training set to classify 3,000 patents into their IPC subclasses. All winning models were based on the Winnow classifier, and triplet features were used as the input [48]. Later, Li et al. [49] proposed a forward ANN-based model and employed the Levenberg–Marquardt algorithm to train the model on a small dataset. Wu et al. [50] proposed a hybrid genetic algorithm with an SVM to classify 234 patents into two sets. Similar to the trend of general document classification, these early machine learning-based studies used manually selected features or statistic-based features as the representation of patents, which may lead to a loss of information.

Several recent deep learning-based approaches have been applied to patent classification research. Grawe et al. [51] proposed an approach that integrates Word2Vec and LSTM to classify patents into 50 categories. Likewise, Shalaby et al. [52] represented patent documents as fixed hierarchy vectors and used an LSTM-based architecture to classify them. Risch et al. [53] proposed domain-specific word embeddings and designed a gated recurrent unit network (GRU)-based model for the patent classification task. Li et al. [12] presented the DeepPatent algorithm based on CNNs and word vectors. Recent studies have also employed transfer learning on large pre-trained language models, including ULMFiT and BERT. Hepburn et al. [54] proposed a patent classification framework based on the SVM and ULMFiT techniques. Kang et al. [55] and Lee et al. [56] fine-tuned the BERT pretrained model to address the patent prior art search task. Abdelgawad et al. [57] applied state-of-the-art hyperparameter optimization techniques to the patent classification problem and presented their effects on the accuracy.

Although deep learning models have achieved a better performance than traditional machine learning-based methods, several limitations remain. First, the performances of current state-of-the-art systems are not sufficiently reliable for real-world large-scale complex technical document management systems [11]. Because existing classification approaches solely use text as the model input and disregard the figures in technical documents, new opportunities exist to improve the classification performance when using multimodal deep learning techniques. Second, all existing approaches are aimed at assigning labels at a single level. To develop a scalable and fine-grained classification system for technical documents, a hierarchical classification is desired [58]. Furthermore, prior studies have used inconsistent datasets and classification schemes for model training and testing, which makes benchmarking and comparisons difficult. A golden standard dataset, such as the patent dataset and the IPC system, may provide a common ground for model training and performance benchmarking of different models.

### B. Multimodal deep learning

Multimodal deep learning aims to design and train models that can utilize, process, and relate information from multiple modalities [59]. Most related reviews claim the superiority of multimodal over unimodal approaches for a series of tasks, including retrieval, matching, and classification [17], [18]. The most common multimodal sources are text, images, videos, and audio. Various multimodal deep architectures have been proposed to leverage the advantages of multiple modalities. For example, Ngiam et al. [60] proposed the learning of shared representations over multiple modalities using multimodal deep learning. Specifically, they concatenated high-level representations and trained two restricted Boltzmann machines (RBMs) to reconstruct the original representations of audio and video. Srivastava and Salakhutdinov [61] proposed a similar approach to modify the feature learning and reconstruction process using deep Boltzmann machines (DBMs).

Furthermore, various neural network architectures have been used to construct multimodal representations [62]–[64]. Each modality starts with some individual neural layers, followed by a specific hidden layer that projects multiple modalities into a joint latent space. The joint representation is then fed into multiple hidden layers or directly followed by a final supervised layer for downstream tasks. Audebert et al. [65] proposed a deep learning-based infused multimodal classifier for documental image classification, utilizing both visual pixel information and the textual content in the images. Their model adopted the MobileNet-v2 model to learn visual features and a simple LSTM network with the FastText representation model [40] to process text data. Despite these promising applications of multimodal deep learning to multimodal data, multimodal technical documents remain mostly unexplored.

### III. METHOD

This section proposes TechDoc, a novel deep learning architecture for multimodal technical document classification. As depicted in Fig. 1, the entire workflow consists of three steps:
1) Data preprocessing;
2) Image and text fusion learning;
3) Network feature fusion learning and document classification.

In the first step, several text preprocessing methods, such as tokenization, phrasing, denoising, lemmatization, and stop-word removal, are applied to convert documents into a suitable representation for the classification model. Compound images are separated into individual ones based on a pre-trained CNN model. A network of the documents based on their associations is constructed. In the second step, image and text feature vectors are jointly trained via neural networks and fused via stepwise concatenation operations. In the third step, the fused features derived from the second step are used as the input document vectors, together with the inter-document association network information, for network fusion learning and final document classification.

Fig. 2 shows the architecture of TechDoc. It consists of three major modules: text feature learning with RNN, image feature learning with CNN, and network feature learning with GNN. The image and text feature learning modules (based on RNN and CNN respectively) are fused first to represent the intrinsic features for individual documents, and then fused with the network learning module (based on GNN) for document classification. In the following, we will describe in detail about these modules and how they are synthesized together in our architecture.

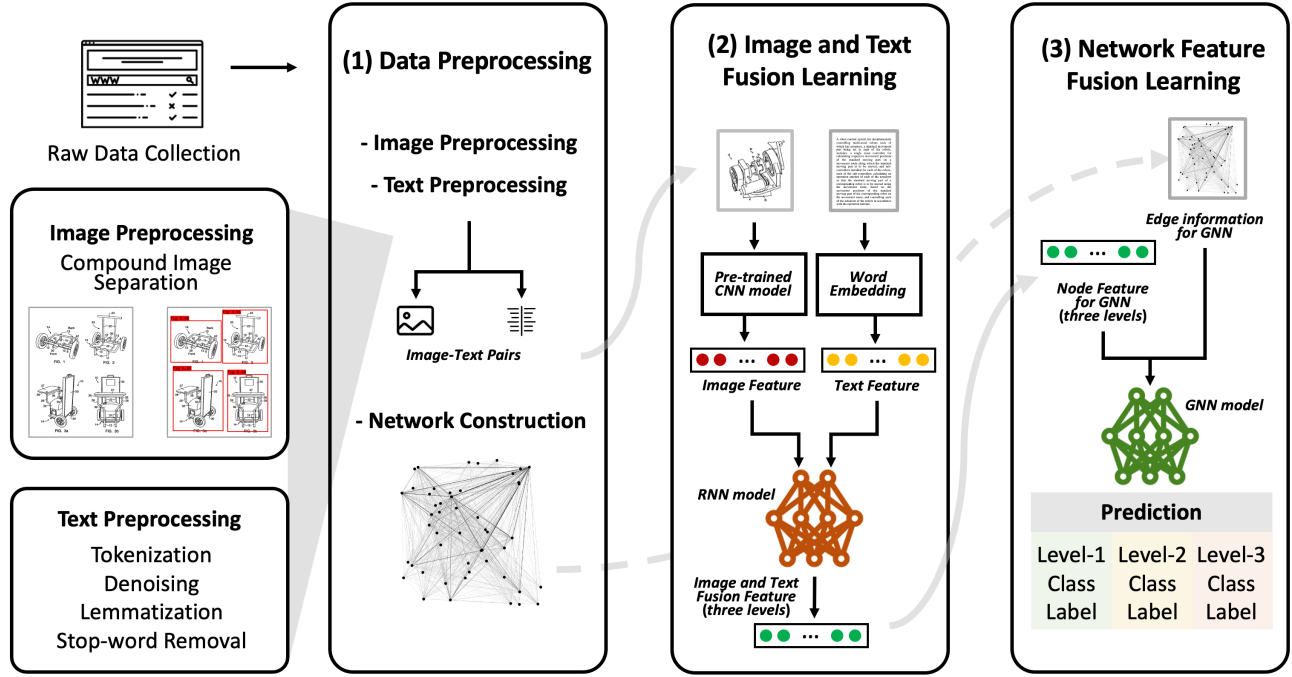

Fig. 1. The entire workflow of using TechDoc to classify technical documents

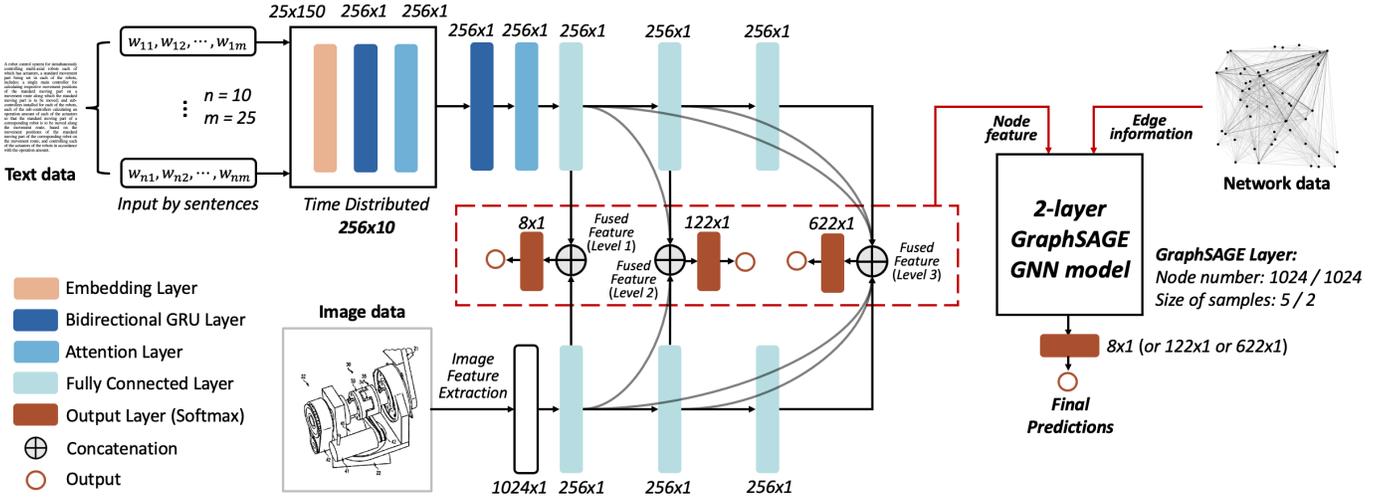

Fig. 2. The architecture of TechDoc based on CNN, RNN and GNN synthesis

*A. Image learning module*

To extract visual features from the technical images, a pre-trained CNN model, VGG-19 [66], is utilized as the image information encoder. It is a robust CNN model and has been widely used in many computer vision applications. It has 19 trainable layers, including convolutional layers, fully connected layers, max-pooling layers, and dropout layers. The VGG-19 model we used is pre-trained on ImageNet [54]. Then, transfer learning techniques are used to fine-tune the image encoder.

Specifically, the final prediction layer of the model is removed and replaced by a new fully connected layer, a dense layer, and an output layer on the top. The modified model is aimed at classifying each image into pre-defined categories to learn corresponding knowledge. Following the training process, the second-to-last fully connected layer of the model is used to extract high-dimensional vectors as image features ($v_{img}$).

*B. Text learning module*

In this part, our model aims to learn word-level, sentence-level, and document-level information from the textual information.

The word encoder is built on the bidirectional recurrent neural network (BRNN) [67], which enables the utilization of the flexible length of contexts before and after the current word position. We used the GRU [68] to track the state of the input sequences without using separate memory cells, which is well suited for extracting long-range dependencies on different time scales. There are two types of gates in the GRU: reset gate $r_t$ and update gate $z_t$. Both aim to control the update of information to the state. At time $t$, the GRU computes its new state as

$$h_t = (1 - z_t) \odot h_{t-1} + z_t \odot \tilde{h}_t \tag{1}$$

This is a linear interpolation between the old state $h_{t-1}$ and the candidate state $\tilde{h}_t$ obtained using the new sequence information. The update gate $z_t$ controls how much previous information will remain and how much new information will be added. Here, $z_t$ is computed as

$$z_t = \sigma(W_z x_t + U_z h_{t-1} + b_z) \tag{2}$$

where $x_t$ indicates the embedding vector with time $t$, and $W$, $U$, and $b$ denote the appropriately sized matrices of the weights and biases, respectively. The symbol $\sigma$ is a sigmoid activation function, and the operator $\odot$ represents an elementwise multiplication. The current state $\tilde{h}_t$ is computed as

$$\tilde{h}_t = \tanh(W_h x_t + r_t \odot (U_h h_{t-1}) + b_h) \tag{3}$$

where the reset gate $r_t$ determines how much information from the old state is added to the current state.

Similar to the unidirectional GRU, the bidirectional GRU processes the input data in two directions with both the forward and backward hidden layers. The computational results of both directions are then concatenated as the output. Let $\overrightarrow{h_t}$ be the forward output of the bidirectional GRU and $\overleftarrow{h_t}$ be the backward output. The final output is the stepwise concatenation of both forward and backward outputs:

$$h_t = [\overrightarrow{h_t}, \overleftarrow{h_t}] \tag{4}$$

Then, the sentence encoder utilizes the word-level representation as the input to build sentence-level vectors using the embedding layer and bidirectional GRU layers. After that, the sentence-level vectors are converted into document-level vectors using different bidirectional GRU layers. Note that not all words and sentences contribute equally to the vector representation. Accordingly, we introduce the attention mechanism [68] to identify essential items for the model.

Assume that the input text has $M$ sentences, and each sentence contain $T_i$ words. Let $w_{it}$ with $t \in [1, T]$ represent the words in sentence $i$. Given a word $w_{it}$, the embedding layer and bidirectional GRU layer convert it into the hidden state $h_{it}$ as

$$\overrightarrow{h_{it}} = \overrightarrow{\text{GRU}}(W_e w_{it}), t \in [1, T] \tag{5}$$

$$\overleftarrow{h_{it}} = \overleftarrow{\text{GRU}}(W_e w_{it}), t \in [T, 1]$$
$$h_{it} = [\overrightarrow{h_{it}}, \overleftarrow{h_{it}}]$$

where $W_e$ indicates the embedding layer matrix, and $\overrightarrow{\text{GRU}}$ and $\overleftarrow{\text{GRU}}$ represent the operations mentioned in the previous section. Then, the attention weights of words $\alpha_{it}$ and sentence vectors $s_i$ can be computed as follows:

$$u_{it} = \tanh(W_w h_{it} + b_w)$$
$$\alpha_{it} = \frac{\exp(u_{it}^T u_w)}{\sum_t \exp(u_{it}^T u_w)} \tag{6}$$
$$s_i = \sum_t \alpha_{it} h_{it}$$

where the context vector $u_w$ can be viewed as a high-level representation of a fixed input over words [69], [70], and is randomly initialized and updated jointly during the training process. Then, another bidirectional GRU layer is used to convert the sentence vectors $s_i$ into hidden state $h_i$ as

$$\overrightarrow{h_i} = \overrightarrow{\text{GRU}}(s_i), i \in [1, M]$$
$$\overleftarrow{h_i} = \overleftarrow{\text{GRU}}(s_i), i \in [M, 1] \tag{7}$$
$$h_i = [\overrightarrow{h_i}, \overleftarrow{h_i}]$$

The attention weights of words $\alpha_i$ and document vectors $v$ can be computed as

$$u_i = \tanh(W_s h_i + b_s)$$
$$\alpha_i = \frac{\exp(u_i^T u_s)}{\sum_t \exp(u_i^T u_s)} \tag{8}$$
$$v = \sum_i \alpha_i h_i$$

where $u_s$ represents the sentence-level context vector, and is randomly initialized and updated, similar to $u_w$.

Through the above training process, the derived document vector $v$ contains hierarchical semantic information from both word-level and sentence-level structures in a technical document. Thus, we call it $v_{txt}$ in the following sections.

### C. Image and text feature fusion learning

In this part, the text feature $v_{txt}$ and image feature $v_{img}$ are fused and jointly trained via several fully connected layers and stepwise concatenation operations. The fully connected and concatenated layers at the end of the model are designed to form a hierarchical learning structure to enable hierarchical classification. For the first-level classification, the fusion process is computed as

$$v_{txt}^{l1} = \sigma(W_{txt}^{l1} v_{txt} + b_{txt}^{l1})$$
$$v_{img}^{l1} = \sigma(W_{img}^{l1} v_{img} + b_{img}^{l1})$$
$$v_{pat}^{l1} = [v_{txt}^{l1}, v_{img}^{l1}] \tag{9}$$
$$p_{l1} = \text{softmax}(W_{pat}^{l1} v_{pat}^{l1} + b_{pat}^{l1})$$

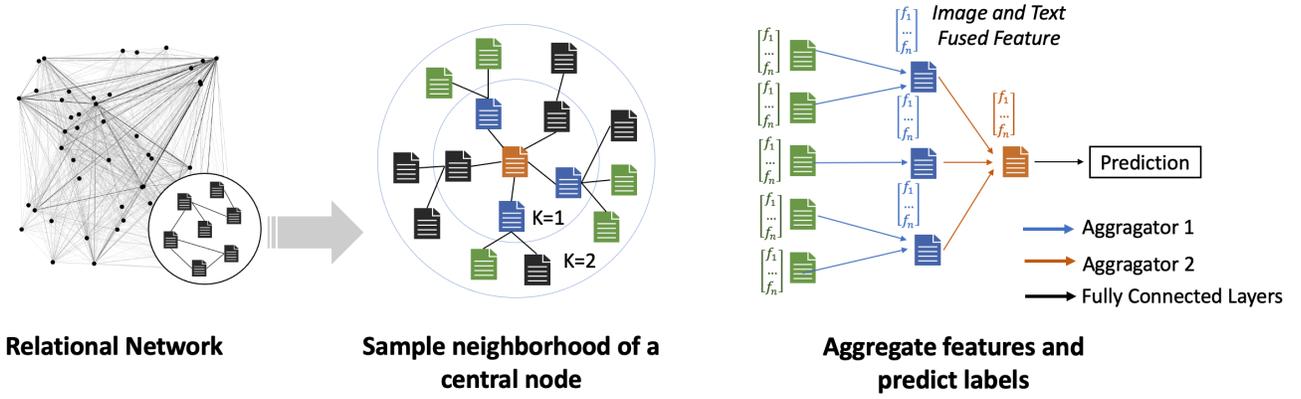

Fig. 3. Illustration of the network fusion learning

where softmax(·) is the softmax activation function, and $p_{l1}$ is the predicted probability vector for the first-level category (not the final classification result). Similarly, the second-level and third-level classification are similar to the above equations. Categorical cross-entropy is used as the training loss: $L = \sum y \log \hat{y}$, where $y$ and $\hat{y}$ denote the predicted label and ground truth, respectively. Because each of the three tasks has an independent loss, the overall loss for our model is

$$L_{overall} = \sum_i \zeta_i M_i, \text{i} = 1,2,3 \quad (10)$$

where $\zeta_i$ is the weight loss, and $\sum_i \zeta_i = 1$. Since lower-level classification is the main task in the case study, and we set 0.05, 0.1, and 0.85 for the three loss weights.

The fully connected layer that aims at the lower-level classification task to learn information from the higher-level task through backpropagation. The numbers of neurons in different layers were tuned for our evaluation case study, and the values are shown in Fig. 2. Finally, we can generate holistic feature vectors that contain both image and text information for individual technical documents from three concatenated layers, corresponding to the hierarchical classification task.

*D. Image, text and network fusion learning*

Training to the step in the previous steps has derived fused feature (containing both image and text information) vectors for individual technical documents. These vectors are then used as node features, together with the network structure of the relations among documents, for the network fusion learning via GNN training. Our GNN is built on the GraphSAGE model [20]. Different from most other GNN methods that are transductive and make predictions on a single fixed graph, GraphSAGE is an inductive technique that allows us to make predictions for unseen nodes, without the need to re-train the embedding model. As illustrated in Fig. 3, the network learning module learns both inter-document relational information and intrinsic document features through aggregating neighboring node attributes. The aggregation parameters are learned by encouraging document pairs co-occurring in short random walks to have similar representations. Then for each document, the network learning module can encode nodes into new vectors considering all three types of information. Finally, a fully connected layer is used to make final label predictions, followed by the GraphSAGE layers. Categorical cross-entropy is used as the training loss. It is noteworthy that we can use the derived node features at three different levels to construct separate GNN models to perform classification tasks at different levels.

IV. EVALUATIVE CASE STUDY

The patent database is widely considered a significant technical document repository because of its large size and holistic classification system of existing technologies and engineering designs [25], [71]–[76]. Patent documents normally contain text, images, citations, and rich information on technologies, products, processes, and systems [21], [31], [77]. In addition, all contents within the database must be both novel (functional and operable) and useful (non-obvious and not proposed earlier). To illustrate the proposed TechDoc model, we first created a benchmark dataset to enable the experiments and comparison among different alternative methods based on the United States Patent and Trademark Office (USPTO) database, which is one of the largest opensource patent datasets. Using the benchmark dataset, we trained the TechDoc model from scratch and utilized the trained model to predict the hierarchical document labels.

*A. Data*

*1) Patent Classification*

All patents contain bibliographic information, including inventors, registration dates, citations, and patent classification codes. There are several types of patent classification schemes, such as the IPC organized by the WIPO, the United States Patent Classification (USPC) organized by the United States Patent and Trademark Office (USPTO), the F-term of Japan, and the Cooperative Patent Classification (CPC) organized by the United States Patent and Trademark Office and the European Patent Office (EPO). Among these classification schemes, IPC is the most widely used.

Approximately 95% of all patents are classified according to the IPC [78].

The IPC scheme is a hierarchical system with increasing levels of resolution corresponding to fine-grained descriptions of technological functionalities [79]. The categories used to classify the documents were organized in a tree-like hierarchy. The basic structure of the IPC uses sections, classes, subclasses, main groups, and subgroups to identify the technologies. Each category in the IPC has its name and code. An example of the hierarchical structure of the IPC code is presented in Table I. Given that the IPC is a well-developed and comprehensive hierarchical structure that covers all thinkable domains of engineering and technology [78], the scope of the IPC classification task meets the requirements of most technology companies to manage their technical documents.

TABLE I
EXAMPLE OF THE IPC HIERARCHICAL STRUCTURE

| IPC Hierarchical Structure | IPC Code | Description |
| --- | --- | --- |
| Section | H | Electricity |
| Class | H01 | Basic electric elements |
| Subclass | H01C | Resistors |
| Main group | H01C 10 | Adjustable resistors |
| Subgroup | H01C 10/02 | Liquid resistors |

In the case study, we focus on the first three levels of the IPC system hierarchy, which correspond to the three-level predictions of our developed method. The system contains 8 sections, 122 classes, and 622 subclasses. The trained model can be used to assign a section label (IPC 1-digit code), class label (IPC 3-digit code), and subclass label (IPC 4-digit code) for any given technical document. Fig. 4 shows illustrative examples of random patents from each IPC section.

Fig. 4. Examples of random patents from each IPC section

*2) Data sampling and curation*

We created a multimodal technical document dataset (containing text and images in the documents and inter-document associations) based on patent data from the USPTO official website[1]. This dataset contains approximately 0.8 million granted patent documents for five years[2]. For every single document, we retain the title and abstract as textual information and independent images separated from the compound images as visual information. We focused on the title and abstract because they are more relevant to the content itself and can ensure the computational efficiency and accuracy of the classification model. By contrast, the claim parts are developed by lawyers using disguised legal languages. The technical description includes a broader background content, which may mislead statistical model training.

Each patent may have more than one IPC code. In our case study, we only focused on the main IPC code, which makes the classification a multi-classification task. It is noteworthy that this benchmark dataset is unbalanced. Fig. 5A shows the percentage of accumulated items against the patent sections, classes, and subclasses. The nonlinearity of the plots demonstrates the imbalance at all three hierarchy levels.

The image separation process, illustrated via Fig. 5B, is described in the next section. Fig. 5C reports the statistical information of the dataset. The codes in the legend represent the section categories (IPC-1 codes). The bars indicate the number of patents per year, and the lines show the number of images before and after separation. In summary, we have collected 798,065 documents with their citations and 13,998,254 independent images.

During the training process, we only used the first image of each document as the visual information. In Table VIII, we also discuss other pre-setting options for the input image number. The results show that using one image per document can lead to the best performance. As the main reason for this phenomenon, it might be that the first image is usually the only image shown on the front page of the patent document and is the most representative among all images. We shuffled and split the dataset as 8:1:1 for training, validation, and testing.

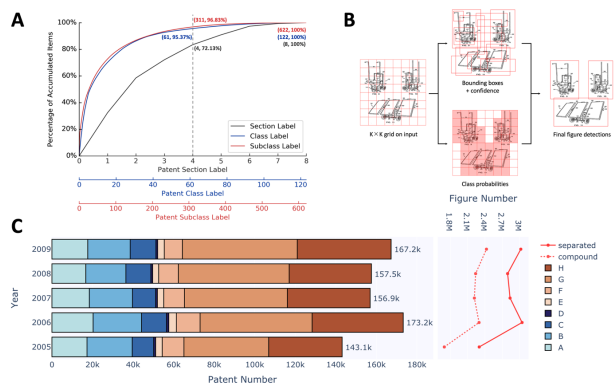

Fig. 5. Data description. A) Percentage of accumulated items against patent category labels of the dataset; B) Illustration of compound image separation based on fine-tuned YOLO-v3 model; C) Statistical information of the dataset by year

---

[1] https://www.uspto.gov/
[2] The USPTO official website has opened US patents with embedded images since March 15, 2001. From 2005, they started using International Common Element (ICE) document type definitions to store the patent data, which allows us to conveniently extract the patent images. Therefore, we use the patents from 2005 to 2009 in our benchmark dataset. In the future, we will expand this dataset to involve more recent patents.

*3) Data preprocessing*

After obtaining raw patent data from the USPTO repository, we observed two problems in our dataset. First, over 30% of patents consist of multiple subfigures [80]. It is inappropriate to feed these compound images as the visual information. Second, the raw text of these documents contains meaningless information, such as stop words and punctuation marks. Therefore, the following preprocessing steps were conducted.

**a) Compound image separation**

In this process, we decomposed compound images into individual figures that can be used as visual input for the classification model. Inspired by literature [80], we leveraged a pre-trained YOLO-v3 convolutional neural network [81] to build an image separator. We followed most of the implementation settings and fine-tuned the network on a manually labeled dataset, which includes 3,600 labeled compound images (3,000 for training and 600 for testing). The separator achieved a 92% accuracy when the threshold value of confidence was set to 0.3, which is an acceptable performance for our data preprocessing. After applying the trained separator to all original images, we obtained 13,998,254 images from 10,877,766 compounded ones.

**b) Text preprocessing**

Words that appear in the patent title and abstract have many structural variants, some of which are not useful for natural language processing. Text preprocessing can reduce the size of the textual dataset and improve the effectiveness and efficiency of the classification model [82]. The text preprocessing includes (1) tokenization, (2) denoising, (3) stop-word removal, and (4) lemmatization.

Tokenization is the process of separating a stream of text into words, symbols, or other elements called tokens, and aims to explore the words in a sentence. Because US patent documents are all in English, most of the words can be separated from each other by white spaces. Next, tokens are standardized and cleaned by a denoising step, which includes converting every term into a lower case and removing numbers, punctuation, and other special characters. The third step is stop-word removal, which aims to drop frequently used stop-words and filler words that add no value to further analysis. We use a widely used list [83] and a USPTO patent stop-word list[3] to identify and remove stop words from the obtained tokens. In the last step, all tokens are converted into their regularized forms to avoid multiple forms of the same word and thus reduce the index size. This operation is achieved by first utilizing a POS tagger [84] to identify the types of tokens in a sentence and accordingly lemmatize them. For example, if the word "studying" is tagged as a VERB by the POS tagger, it would be converted into "study," but remain as "studying" when tagged as a NOUN. All text preprocessing steps were applied using the natural language toolkit (NLTK) [85], which is a suite of text processing libraries using the Python programming language.

**c) Image and text feature extraction**

[3] http://patft.uspto.gov/netahtml/PTO/help/stopword.htm

As described in Section II, we use a VGG19 network as the image encoder [66]. Each individual technical image is represented as a 1,024-dimension vector. We utilize class activation mapping (CAM) technique [86] to highlight the importance of the image region for the neural network. Fig. 6 illustrates the informative regions of five random patent images.

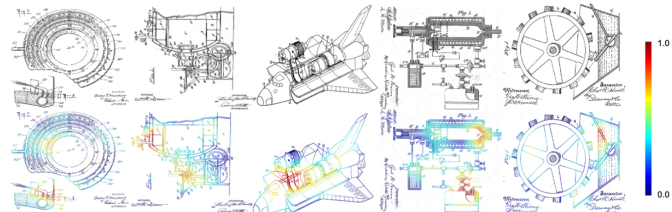

Fig. 6. Informative regions of five random patent images for the image encoder identified via CAM technique [86]

To encode the textual information, we used the TechNet pre-trained word embedding vectors to represent every single token. TechNet is a semantic network consisting of words and phrases contained in patent titles and abstracts from the USPTO patent database, and also provides embedding vectors for technical terms [87]. Using TechNet, each word is converted into a 150-dimension vector as the input of the multimodal deep learning model.

*B. Experiment setup*

To evaluate our proposed TechDoc trimodal deep learning model (*text+image+network*), we first compared it against three unimodal models (*text-only*, *image-only*, *network-only*) and two dual-modal model (*text+image*, *text+network*). All these models are originated from one or more modules of the total TechDoc architecture and trained with the same experimental settings. For example, in the *text-only* unimodal model, the input data have original values for text and zero values for images. As for the *network-only* model, we adopt the basic majority voting strategy based on the forward citations of any given patent to make predications. Moreover, we implemented five different patent classification algorithms [12], [52], [53], [56], [88], three general document classification algorithms [40]–[42] and a multimodal classification model [65] on the same dataset for comparison with TechDoc.

One challenge in training TechDoc is tuning the pre-setting hyperparameters. Table II lists the hyperparameter space and the range of potential values or settings. We used a benchmark method to randomly search in the hyperparameter space to identify the configuration that leads to a high performance [89]. In our study, 50 neural network models with different configurations were trained. The final selected settings for the text processing are shown in Table II, which represents the best combination in the potential hyperparameter setting space. For the neural network parameters, we set the dimension of the GRU to 128. In this case, as a combination of forward and backward GRUs, the dimensions of both word and sentence feature vectors are 256. We set the dimension of the fully connected layer to 256 with He's uniform initialization [90]. As for the training process, we specified the batch size to 64. We set 25 as the maximum number of sentences in a document, and 10 as the maximum number of words in a sentence. In the image and text fusion learning module, we applied the Adam optimizer [91] with the best learning rate using a grid search on the validation set. The iteration time was set to 10, which was sufficient for the convergence of the model's loss function. In the network learning module, we built a two-layer GraphSAGE model with 1,024 nodes in each layer. As for the number of sampled neighbors, we set the sizes of 1-hop and 2-hop neighbor samples to be 5 and 2. The node numbers of the last several fully connected layers are corresponding to the specific tasks (8 or 122 or 622). We stacked the GraphSAGE layers and fully connected layer in the model, and defined the category cross-entropy as the loss function. Adam optimizer was used again for the GNN training. We set the iteration time of GNN training to 10. The aggregate functions and other parameters were set as suggested in the original GraphSAGE paper [20]. All experiments were conducted on a machine with an Nvidia Titan X 16 GB GPU and 64 GB of RAM.

TABLE II
PRE-SETTING HYPERPARAMETER SPACE

| Pre-setting item | Range | Selected setting |
|---|---|---|
| Input text constitution | • Title only<br>• Abstract only<br>• Title+Abstract<br>• Title×3+Abstract | Title+Abstract |
| Word Embedding | • TechNet [87]<br>• Glove-300D [92]<br>• Glove-50D [92]<br>• ConceptNet [93]<br>• Wikipedia2Vec [94] | TechNet |
| Number of input words (First K words) | [50, 100, 150, 200, 250, 300, 350, 400] | 250 |

We used the top-1, top-5, and top-10 accuracies and reciprocal average rank (RAR) measures to evaluate performances of different models. The top-K accuracy calculates the percentage of correct labels within the top-K-predicted scores. The RAR measures how far down the ranking of the correct label is. It can be calculated as follows:

$$RAR = \frac{1}{1/N \times \sum_{i=1}^{N} rank(y_j)} \quad (11)$$

where $N$ indicates the number of documents, and $rank(y_j)$ represents the ranking position of the ground-truth label in the predicted score list.

### C. Experimental results

To understand which modality is more critical for document classification, we conducted an ablation study to analyze the experimental results on the benchmark patent dataset for six models, including three unimodal models (*image-only, text-only, network-only*), two dual-modal models (*text+image, text+network*) and the tri-modal TechDoc model (*text+image+network*). All the models were run 10 times. The results (mean±standard deviations) of all metrics are reported in Table III, IV, and V. Firstly, we can see that the TechDoc model, which fuses text, image and network information together, outperforms other models on all tasks and metrics significantly based on student t-test (p < 0.05) [95]. Looking at three unimodal models, we can find that using the *text-only* or the *network-only* models can get reasonably good performances, which are much better than performances of the *image-only* model. In addition, both of the two dual-modal models outperform all three unimodal models. It is not surprising to see that removing the network learning module makes more impact on the model performance than removing the image module. These findings reveal that the text and network information of technical documents are more important for classification while involving technical visual information can additionally bring a modest and consistent advantage for the model.

TABLE III
SECTION (IPC 1-DIGIT) CLASSIFICATION RESULTS

| Model | Top-1 Acc. | Top-5 Acc. | Top-10 Acc. | RAR |
|---|---|---|---|---|
| Image-only | 0.493±0.002 | 0.965±0.001 | - | 0.490±0.001 |
| Text-only | 0.768±0.001 | 0.995±0.000 | - | 0.742±0.001 |
| Network-only | 0.779±0.000 | 0.930±0.000 | - | - |
| Text+Image | 0.783±0.001 | 0.995±0.001 | - | 0.754±0.001 |
| Text+Network | 0.822±0.002 | 0.998±0.001 | - | 0.789±0.001 |
| **Image+Text+Network (TechDoc)** | **0.825±0.003** | **0.998±0.001** | - | **0.791±0.002** |

TABLE IV
CLASS (IPC 3-DIGIT) CLASSIFICATION RESULTS

| Model | Top-1 Acc. | Top-5 Acc. | Top-10 Acc. | RAR |
|---|---|---|---|---|
| Image-only | 0.341±0.003 | 0.657±0.001 | 0.780±0.004 | 0.114±0.003 |
| Text-only | 0.691±0.003 | 0.937±0.003 | 0.973±0.001 | 0.450±0.002 |
| Network-only | 0.681±0.000 | 0.877±0.000 | 0.883±0.000 | - |
| Text+Image | 0.713±0.001 | 0.946±0.002 | 0.976±0.001 | 0.460±0.001 |
| Text+Network | 0.720±0.002 | 0.946±0.001 | 0.976±0.003 | 0.464±0.001 |
| **Image+Text+Network (TechDoc)** | **0.724±0.004** | **0.948±0.002** | **0.977±0.004** | **0.471±0.003** |

TABLE V
SUBCLASS (IPC 4-DIGIT) CLASSIFICATION RESULTS

| Model | Top-1 Acc. | Top-5 Acc. | Top-10 Acc. | RAR |
|---|---|---|---|---|
| Image-only | 0.245±0.002 | 0.471±0.003 | 0.581±0.004 | 0.031±0.002 |
| Text-only | 0.618±0.005 | 0.886±0.002 | 0.936±0.003 | 0.233±0.004 |
| Network-only | 0.587±0.000 | 0.809±0.000 | 0.826±0.000 | - |
| Text+Image | 0.636±0.002 | 0.896±0.002 | 0.944±0.001 | 0.236±0.002 |
| Text+Network | 0.649±0.003 | 0.908±0.002 | 0.955±0.003 | 0.281±0.002 |
| **Image+Text+Network (TechDoc)** | **0.655±0.001** | **0.916±0.001** | **0.960±0.003** | **0.292±0.001** |

TechDoc was then compared with nine prior relevant deep learning models, including five patent document classification models, three general text classification models and one multimodal document classification model, on the same dataset. All used titles and abstracts as textual inputs and aimed at predicting the IPC 4-digit subclass labels. Specifically, for the fine-tuned BERT model, we leveraged the released BERT-Base pre-trained model (Uncased: 12-layer, 768-hidden, 12-heads, 110 million parameters) [96], as the author claimed in the literature [56]. Table VI shows the performance of each model. TechDoc outperforms baselines significantly based on *t*-test (*p* < 0.05) for all indicators.

There might be three reasons why our model shows better performances than others. First, compared to the models that only focus on processing text data, adding information of image and network can bring additional predictive power (also shown in Table III, IV, and V). Second, compared to the other dual-modal method (model (i)) that uses both text and image information for classification, the dual-modal model based on the TechDoc architecture (model (j)) also shows its advantages on performance. This is because the TechDoc is explicitly designed for the technical document classification. In contrast, the model (j) is designed for documental image classification by utilizing visual pixel information and the textual content in the images. Third, our model aims to classify text documents into a given hierarchy, which enables the model to share technical knowledge at different levels, whereas other methods regard the predictions at different levels as separate tasks.

TABLE VI
SUBCLASS (IPC 4-DIGIT) CLASSIFICATION RESULTS WITH DIFFERENT MODELS

|     | Model | Top-1 Acc. | Top-5 Acc. | Top-10 Acc. | RAR | Training time (minutes) |
| --- | --- | --- | --- | --- | --- | --- |
| (a) | DeepPatent [12] | 0.615±0.003 | 0.885±0.010 | 0.940±0.002 | 0.233±0.004 | 9.7 |
| (b) | LSTM + FHV [52] | 0.605±0.009 | 0.878±0.007 | 0.929±0.007 | 0.221±0.005 | 14.3 |
| (c) | LSTM + Word2Vec [88] | 0.591±0.008 | 0.865±0.004 | 0.922±0.002 | 0.206±0.004 | 13.5 |
| (d) | GRU + Word2Vec [53] | 0.617±0.008 | 0.884±0.005 | 0.934±0.005 | 0.233±0.004 | 13.0 |
| (e) | Fine-tuned BERT [56] | 0.624±0.002 | 0.891±0.002 | 0.943±0.002 | 0.235±0.002 | 1,290.3 |
| (f) | TextRCNN [41] | 0.586±0.004 | 0.860±0.005 | 0.901±0.005 | 0.181±0.005 | 15.7 |
| (g) | HAN [42] | 0.606±0.001 | 0.879±0.001 | 0.932±0.001 | 0.214±0.001 | 17.7 |
| (h) | FastText [40] | 0.484±0.003 | 0.797±0.002 | 0.846±0.001 | 0.100±0.001 | 3.7 |
| (i) | Audebert et al. [65] | 0.537±0.002 | 0.818±0.002 | 0.875±0.001 | 0.135±0.002 | 63.5 |
| (j) | This work (Text+Image) | 0.636±0.002 | 0.896±0.002 | 0.944±0.001 | 0.236±0.002 | 77.0 |
| (k) | This work (Text+Network) | 0.649±0.003 | 0.908±0.002 | 0.955±0.003 | 0.281±0.002 | 36.3 |
| (l) | TechDoc (Image+Text+Network) | **0.655±0.001** | **0.916±0.001** | **0.960±0.003** | **0.292±0.001** | 95.3 |

To evaluate the computing efficiency of different models, we report the training time (10 epochs for each model) in Table VI. We can see that the TechDoc consumes more training time (model (l), 95.3 minutes) than the unimodal methods that only process texts, but significantly less time than the second most accurate model, Fine-tuned BERT (model (e), 1290.3 minutes). It is important to note that, multimodal data fusion would naturally require extra computation and training time than unimodal learning [97]. In our case, processing and fusing additional image and network information to texts is expected to incur extra computations and training time. We can also find the dual-modal method that fuses text and network information based on TechDoc architecture can reduce much training time and achieve a better performance than other baselines. When training efficiency is the top priority for users, they can choose the dual-modal method (*text+network*) to build their own document classifier.

We then conducted a time-complexity analysis [98] on the training time increase for unit accuracy improvement of the models, using the fastest model FastText as the baseline. $\Delta t$ denotes the difference between the training times of a model and FastText. $\Delta a$ denotes the difference between the Top-1 accuracy values of the model and FastText. Then, $\Delta t/\Delta a$ indicates how much additional training time the model would require to obtain unit accuracy increase. As reported in Table VII, our model took additional 5.4 minutes to improve each 1% of top-1 accuracy from FastText, which is better than another multimodal model (model (i)) and the second most accurate model (model (e)). When using DeepPatent as the baseline, our model took additional 21.4 minutes to improve each 1% of top-1 accuracy.

TABLE VII
TIME-COMPLEXITY ANALYSIS WITH DIFFERENT MODELS

| Model | $\Delta t$ (minute) | $\Delta a$ (%) | $\Delta t/\Delta a$ (minute/%) |
| --- | --- | --- | --- |
| DeepPatent [12] | 6.0 | 13.1 | 0.5 |
| LSTM + FHV [52] | 10.6 | 12.1 | 0.9 |
| LSTM + Word2Vec [88] | 9.8 | 10.7 | 0.9 |
| GRU + Word2Vec [53] | 9.3 | 13.3 | 0.7 |
| Fine-tuned BERT [56] | 1286.6 | 14.0 | 91.9 |
| TextRCNN [41] | 12.0 | 10.2 | 1.2 |
| HAN [42] | 14.0 | 12.2 | 1.1 |
| Audebert et al. [65] | 59.8 | 5.3 | 11.3 |
| This work (Text+Image) | 73.3 | 15.2 | 4.8 |
| This work (Text+Network) | 32.6 | 16.5 | 2.0 |
| TechDoc (Image+Text+Network) | 91.6 | 17.1 | 5.4 |

In sum, our model presents statistically significant accuracy advantages over all prior models, despite requiring additional training time to process and fuse additional image and network information. A user may employ the TechDoc, when classification performance is the top priority, and the training time is affordable for the specific user. In the case study, training TechDoc took 95 minutes for around 0.6 million documents with only one GPU. In real-world applications with a much larger training dataset, users may consider training TechDoc on more powerful computing infrastructures, such as GPU clusters and cloud-computing platforms, to limit training time.

### D. Comparison of unimodal and multimodal models

Fig. 7 reports the confusion matrices of three unimodal models and the multimodal model (i.e., TechDoc) for patent section classification[4]. These matrices were built based on the predicted results of the test set. In Fig. 7, we can observe that TechDoc improves the classification of sections C, D, E, F, G and H when compared to the best performance of the unimodal models. The increase in accuracy ranges 0.02 to 0.08. The *text-only* model outperforms the *image-only* model in every section, reflecting the natural difficulty in classifying patents using only images even for human experts. For example, some technical images only present a partial view of a specific design, which cannot represent the entire product. In Fig. 7a and 7b, we can find both models achieve a better accuracy in sections A, B, G, and H than in the other sections. The imbalanced patent dataset has a more significant impact on the *image-only* model than others. The *image-only* model even ignores section D, which is quite a small group. In Fig.

---

[4] The brief meaning of eight sections (Section A-H) is shown in Fig. 4. The detailed descriptions of eight sections can be viewed at the WIPO website: https://www.wipo.int/classifications/ipc/ipcpub

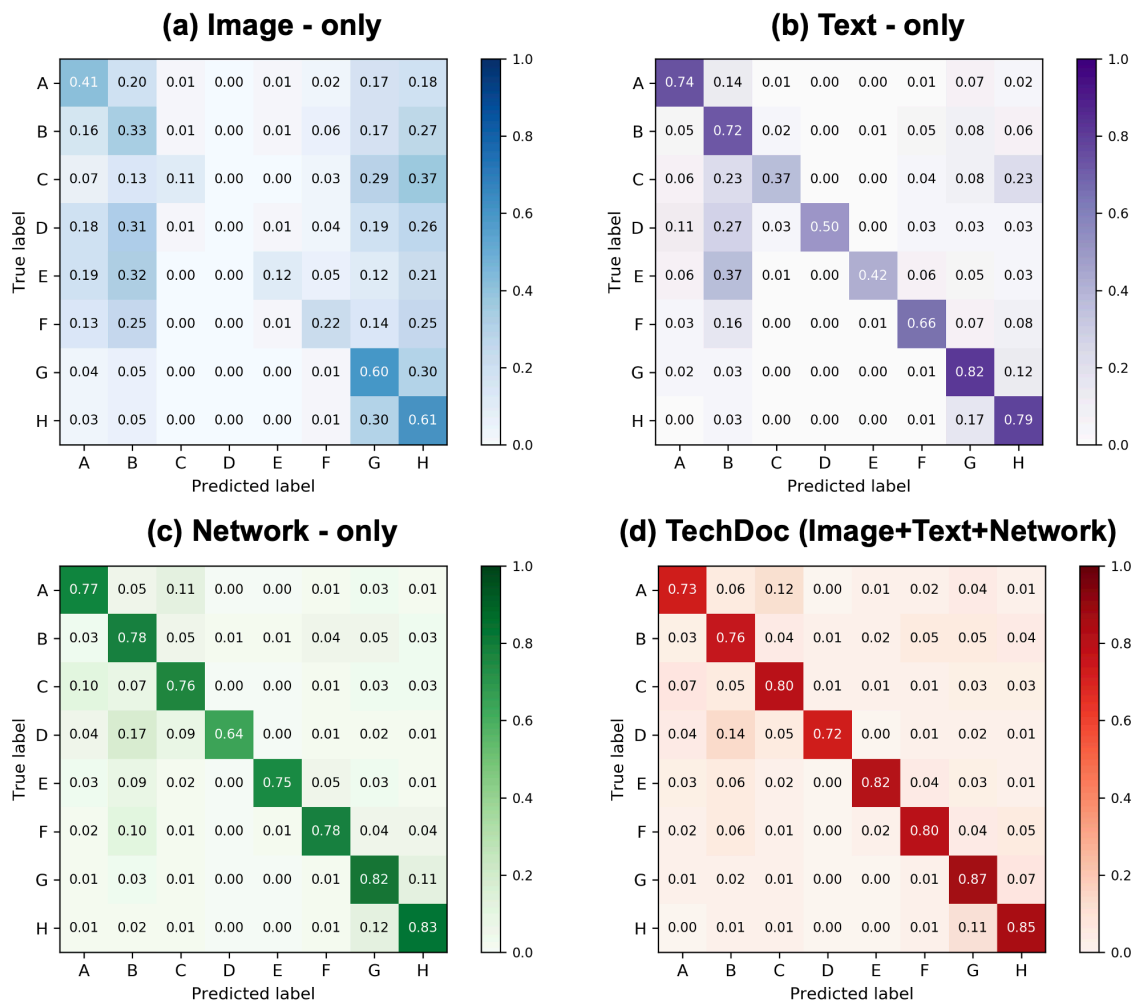

Fig. 7. Confusion matrices of four models on the patent section (IPC 1-digit) classification task

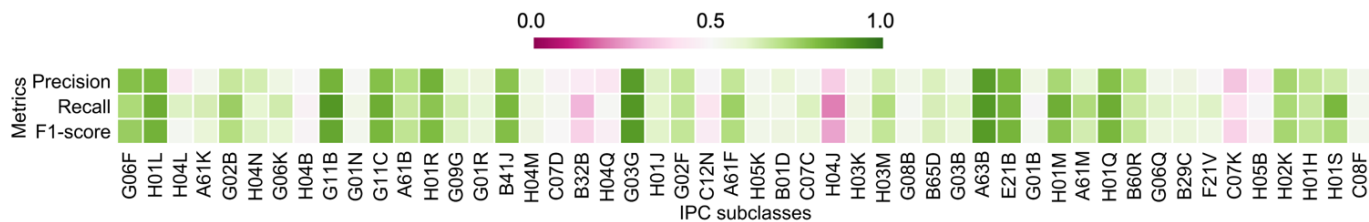

Fig. 8. Categorial classification performance (precision, recall and f1-score) using TechDoc for the biggest 50 subclasses

7c, we can find that the *network-only* model has a good performance for small groups, which shows the value of involving relational information. TechDoc (Fig. 7d) outperforms all the unimodal models in section D. This may be because TechDoc can learn some internal relationships among three modalities, and the superposition of information is nonlinear for the model. Moreover, in the second column of Fig. 7b and 7c, we can see that most misclassifications occur in Section B. The same situation is also shown in Fig. 7d, whereas TechDoc alleviates this problem to a certain extent.

*E. Analysis of the performance across subclasses*

In this section, we move forward to explore the classification performance across subclasses. We selected the biggest 50 subclasses based on their size (number of patents) and computed their precision, recall, and F1-score. The results are presented in Fig. 8. The subclasses are in descending order from left to right according to size[5].

Fig. 8 shows that the performances of 50 subclasses are very uneven. We should note that an imbalanced dataset might lead to a bad performance for some small groups. Although the F1-score ranges from 0.27 to 0.88, most of the subclasses (43/50) achieve a score of higher than 0.50. This result and the overall trends of three curves reveal that there is only a modest correlation between performance and size for these 50 subclasses.

---

[5] The detailed descriptions of the 50 subclasses can be viewed at the WIPO website: https://www.wipo.int/classifications/ipc/ipcpub

Among all subclasses, eight groups have an F1-score of higher than 0.8. The best five groups are: *G03G* (Electrography, electrophotography and magnetography), *A63B* (Apparatus for physical training), *G11B* (Information storage based on relative movement between record carrier and transducer), *H01L* (Semiconductor devices), *E21B* (Earth or rock drilling apparatus). By checking the brief descriptions of these subclasses, we find that most of them are related to devices, apparatus, or specific technical objects. Patents from these groups usually contain distinct domain-related information from both text and images, which may at least partially explain the better accuracy.

In addition, we find three subclasses whose F1-score is lower than 0.4: *B32B* (layered products), *C07K* (peptides), and *H04J* (multiplex communication). For the first subclass, *B32B*, the reason may be the abstract description of the domains themselves. Patent documents that belong to layered products can come from entirely different disciplines. According to official documents from the WIPO[6], both layered honeycomb and layered cellular are covered by this subclass. For the other two subclasses, *C07K* and *H04J*, some similar categories also exist. For example, our TechDoc model misclassified a patent in the test set named "transmission line monitoring system" (US12491709) into the subclass *H04J*. The ground-truth label of this patent is *H04B* (transmission), which shares similar technical content of *H04J*. Thus, when predicting such types of subclasses, the top-5 returned labels from our model can more meaningful.

### F. Discussion on the influence of data preprocessing

As described in previous section, the application of the proposed model requires a series of automatic data preprocessing steps, which may introduce biases. For the image preprocessing, we leveraged a fine-tuned CNN model to separate compound images with an accuracy of 92%, which means there still exists a few compound images used as the input of the visual end during the model training. A cleaner image dataset would further bring some improvements to the current performance of our proposed model. In addition, the text-image pairs used in our experiments only contain one image (the image shown on the first page) per document. We tested some alternative settings (using five or all images per document as input images), and found using one image per document achieves the best result (Table VIII). There are several reasons. First and foremost, the first image (usually shown on the front page of the patent) is often the most important and representative one to a patent. Second, a patent may have several similar images, which contain redundant information to the model. Third, some types of images do not provide much technical-related visual information, such as flowcharts and tables. Involving these kinds of images may even harm the performance of our model. We should note that the method we use to combine all images in the experiments is a simple way of the early fusion. There is still room for future studies to explore the way to utilize all images of technical documents.

TABLE VIII
SUBCLASS (IPC 4-DIGIT) CLASSIFICATION RESULT UNDER DIFFERENT NUMBER OF INPUT IMAGES

| # of input images | Top-1 Acc. (%) | Top-5 Acc. (%) | Top-10 Acc. (%) | RAR |
|---|---|---|---|---|
| 1 image per doc. | **0.655±0.001** | **0.916±0.001** | **0.960±0.003** | **0.292±0.001** |
| 5 images per doc. | 0.654±0.000 | **0.916±0.002** | 0.955±0.001 | 0.290±0.001 |
| all images in doc. | 0.631±0.003 | 0.907±0.004 | 0.949±0.001 | 0.234±0.002 |

For the text preprocessing, we applied the general steps to clean the original technical text for further model training, including tokenization, denoising, stop-word removal, and lemmatization. It is noteworthy that all these automatic preprocessing steps may involve biases in real-world applications. Prior fundamental research [99] studied the different combinations of preprocessing methods and their influence on the model performance, and pointed out that the best text preprocessing strategy might be different for different datasets. In the future work, we plan to conduct a more detailed experiment to find out the best text preprocessing strategy for technical document classification.

## V. DISCUSSION ON APPLICATIONS OF THE TECHDOC

Thus far, we have presented a deep learning model for hierarchical classification of multimodal technical documents, i.e., TechDoc, which combines three types of information in training, and demonstrated and tested it with patent document classification. The application of TechDoc is not limited to patent data and IPC. In practice, it can be also trained on non-patent technical documents for technological labels at different levels of a well-defined hierarchy. TechDoc can be used to help companies automatically classify and manage their technical documents, particularly newly generated ones, in several practical ways.

Table IX shows the taxonomy of neural network training strategies. First of all, companies that already have sufficient technical documents classified in their own document categorization system (which most large established companies have) can use their classified/labeled document data to train a model based on the TechDoc architecture and the workflow we introduced above and apply the trained model to automatically classify documents generated later. This is strategy #1 in the taxonomy in Table IX. Alternatively, they may directly use the model trained on patent data and IPC to automatically classify documents generated later, without requiring new categorical labels, i.e., strategy #2 in Table IX.

---
[6]https://www.wipo.int/classifications/ipc/en/

TABLE IX
TAXONOMY OF NEURAL NETWORK TRAINING STRATEGIES

| Document Categories | | Proprietary Database for Training | |
|---|---|---|---|
| | | Large | Small |
| | Proprietary | 1) Train a network with proprietary data and own categories | 3) Pre-train a network with the patent database and IPC and then re-train/fine-tune it with relatively smaller proprietary dataset and own categories |
| | IPC | 2) Train a network with the patent database and IPC | 2) Train a network with the patent database and IPC |

Some companies (e.g., small enterprises or new startups) might not have many documents themselves to train the TechDoc model from scratch. As illustrated in the case study, TechDoc can be trained using the multimodal engineering dataset that we created based on patent data. The trained model can automatically classify newly generated non-patent documents into respective categories in the IPC hierarchy. This is again strategy #2 in Table IX. Furthermore, the trained model can be used to extract the features of multimodal documents from the hidden layers. Using unsupervised clustering methods, such features enable the document owners to identify clusters or categories that are different from the initial predefined patent classification scheme, and guide them to define their own categories.

For those companies preferring their own document categorization system but having insufficient in-house labelled documents for high-performance neural network training, they may consider a transfer learning strategy. That is to first utilize the large patent database to pre-train a neural network for the IPC task (like the one we trained in this research), and then further re-train/fine-tune the network with the relatively smaller set of proprietary documents and their categorical labels. This is strategy #3 in Table IX. For the transfer, one may freeze the parameters of the pre-trained network, remove the topmost layers, add several layers on the top including the final fully-connected output layer with the same dimensions of the in-house categorization system, and randomly initialize the new parameters added to the structure, for re-training with the proprietary data and categorical labels.

In addition to the above scenarios, the TechDoc architecture can also be applied for classification when documents only involve one or two types of information. The results of the ablation study (Table III, IV, and V) show that using the *text-only* or the *network-only* model can obtain reasonably good performances. Compared to alternative methods, Table VI shows the two dual-modal models based on the TechDoc architecture (*text+image, text+network*) and the tri-modal TechDoc model (*text+image+network*) can outperform other existing models. This finding suggests that our model is more suitable and competitive for classifying multimodal engineering documents comprising two or three modalities. It is noteworthy that the training efficiency of the dual-modal (*text+network*) model is better than the tri-modal TechDoc model, according to Table VI. In this case, when training efficiency is the top priority for users, they can choose the dual-modal model (*text+network*) to build their own document classifier. When classification performance is the top priority and the training time is affordable for the specific users, they can employ the TechDoc model that combines three types of information.

## VI. CONCLUDING REMARKS

The engineering design, analysis, and manufacturing processes generate many diverse technical documents that describe technologies, products, processes, and systems. For large engineering companies, the number of technical documents that need to be managed and organized for retrieval and reuse has grown dramatically and demands more scalable and accurate document classification. Therefore, automated approaches increase their potential value in reducing the burden of experts and supporting diverse analytical reasons for classification. Herein, we propose a multimodal deep learning architecture (i.e., TechDoc) for technical document classification that can take advantage of three types of information (images and texts of documents, and relational network among documents) and assign documents into hierarchical categories.

TechDoc synthesizes the CNN, RNN and GNN through an integrated training process. To illustrate the proposed method, we applied it to a large multimodal technical document database of about 0.8 million patents and trained the model to classify technical documents based on the hierarchical IPC system. We demonstrated that the multimodal fusion model outperforms the unimodal models and baseline models significantly. There is still much room for improvement. First, how to identify an effective way to utilize the information in document images is still unclear. In this study, we only utilized the first images of patents and leveraged a fine-tuned CNN network to extract visual features. Determining how to take advantage of all patent images remains a challenge. Second, some patents have more than one IPC code, which makes such a classification task a multilabel classification problem. Although some prior research has presented certain achievements [12], [52], [56], it remains challenging to determine the exact number of categories. Third, the current training workflow is computing-intensive because of the large number of free weights in the multimodal deep learning model. In the future, we plan to further improve the model by exploring alternative and more efficient ways to mine multimodal information, especially visual information. Besides, further research can also explore whether other types of technical document datasets (especially the technical documents generated in large engineering companies) and other technical-related classification systems (e.g., USPC and CPC) are more generally amenable to higher accuracies for the TechDoc model. In addition to classification, some prior studies have also utilized text mining techniques to automatically analyze and manage technical documents, including topic modeling [100], [101], and subject-action-object semantic structure extraction [102], [103]. Researchers may combine these AI techniques, in conjunction with document classification methods, to develop more powerful technology management systems for real-world applications.

ACKNOWLEDGMENT

The authors acknowledge the funding support for this work received from the SUTD-MIT International Design Center and SUTD Data-Driven Innovation Laboratory (DDI, https://ddi.sutd.edu.sg/), National Natural Science Foundation of China (52035007, 51975360), Special Program for Innovation Method of the Ministry of Science and Technology, China (2018IM020100), National Social Science Foundation of China (17ZDA020), and the China Scholarship Council. Any ideas, results and conclusions contained in this work are those of the authors, and do not reflect the views of the sponsors.

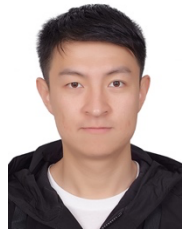

**Shuo Jiang** is a Ph.D. student in the School of Mechanical Engineering at Shanghai Jiao Tong University (SJTU). He had been a visiting Ph.D. student at Institute for Data, Systems, and Society (Design and Invention Group) at Massachusetts Institute of Technology (MIT) for one year sponsored by the Chinese Scholarship Council, and also a visiting Ph.D. student at Data-Driven Innovation Lab at Singapore University of Technology and Design (SUTD). He received a B.S. degree in the School of Mechanical Engineering at East China University of Science and Technology (ECUST). His research interests include data-driven design, machine learning-based engineering design, and computational design methods.

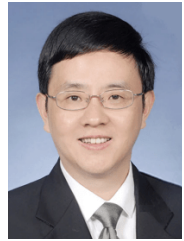

**Jie Hu** received the Ph.D. degree from Zhejiang University, Hangzhou, China, in 2001. He is currently a tenured Full Professor in the School of Mechanical Engineering at Shanghai Jiao Tong University, Shanghai, China. Prior to joining SJTU, he was a post-doctor at Tsinghua University. His research interests include innovative design, design theory, artificial intelligence, and computer-aided design.

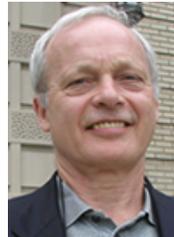

**Christopher L. Magee** is a professor in the engineering systems division and mechanical engineering department, MIT, and codirector, International Design Center, associated with the Singapore University of Technology and Design being codeveloped by MIT and Singapore. His recent research has emphasized innovation and technology development in complex systems. He was elected to the National Academy of Engineering (1997) while with the Ford Motor Company for contributions to advanced vehicle development. He was a Ford technical fellow (1996), and is a fellow of the American Society for Metals. He has a B.S. degree and M.S. and Ph.D. degrees in metallurgy and materials science, respectively, from Carnegie Institute of Technology (now Carnegie Mellon University), and an M.B.A. degree from Michigan State University.

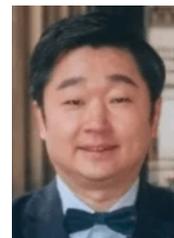

**Jianxi Luo** received the B.E. and M.S. degrees in engineering from Tsinghua University, Beijing, China, in 2001 and 2004, respectively, and the S.M. degree in technology policy and the Ph.D. degree in engineering systems (technology management and policy track) from the Massachusetts Institute of Technology, Cambridge, MA, USA, 2006 and 2010, respectively. He is the Founder and the Director of Data-Driven Innovation Lab, Singapore University of Technology and Design (SUTD), Singapore. He was the Director of SUTD Technology Entrepreneurship Programme. He teaches topics on engineering entrepreneurship, design, and innovation. His research interests include data-driven innovation and artificial intelligence for design. He is the Department Editor of the IEEE TRANSACTIONS ON ENGI- NEERING MANAGEMENT, an Associate Editor for *Design Science* and *Artificial Intelligence for Engineering Design, Analysis and Manufacturing*, an Editorial Board Member of *Research in Engineering Design*, etc. He was the Chair of INFORMS Technology Innovation Management and Entrepreneurship Section